\documentclass[11pt]{article}
\usepackage[margin=1in]{geometry}
\usepackage[T1]{fontenc}
\usepackage{lmodern}
\usepackage{microtype}
\usepackage{amsmath,amssymb}
\usepackage{graphicx}
\usepackage{booktabs}
\usepackage{enumitem}
\usepackage[numbers,sort&compress]{natbib}
\usepackage[colorlinks=true,linkcolor=blue,citecolor=blue,urlcolor=blue]{hyperref}
\usepackage{url}
\setlist{itemsep=2pt,topsep=4pt}
\newcommand{\yes}{$\checkmark$}\newcommand{\some}{$\circ$}

\title{MoE-Context: Show the Model Only What It Needs}
\author{Vedant Navale\\ UpThink EduTech Services Pvt.\ Ltd.}
\date{November 2026}

\begin{document}
\maketitle

\begin{abstract}
AI agents read their whole history at every step. People do not work like that: we keep what the moment needs and hold the rest as a rough note we can recall. MoE-Context applies this to language models. The context is split into blocks. Before each step a fast yes or no pass decides which blocks the step needs; the rest stay as one-line notes that can be brought back, and nothing is deleted. In the version we build, one model makes this decision and then answers, over a single key-value cache where hidden blocks are masked rather than removed. On a laptop with Qwen3-1.7B and a synthetic lookup task, showing only the needed block raises accuracy from 93 to 119 of 120, and a masked block is as hidden as a deleted one. The model's own picking is the weak part. A supervised linear probe written into spare output rows lifts its accuracy from 80 to 110 of 120, but simple keyword search picks perfectly on this task, and the probe does not transfer to a second task. We report these results openly, along with two cheap usage patterns that block switching enables: undoing a failed attempt and running subagents without a new context.
\end{abstract}

\section{Introduction}

You do not load your whole life into your head before making tea. You hold what the task needs; older things stay as a faint memory, and with a little effort the detail comes back. An agent today does the opposite: every message, every reply and every file it has read sits in its context at every step.

\begin{figure}[htbp]
\centering
\includegraphics[width=\linewidth]{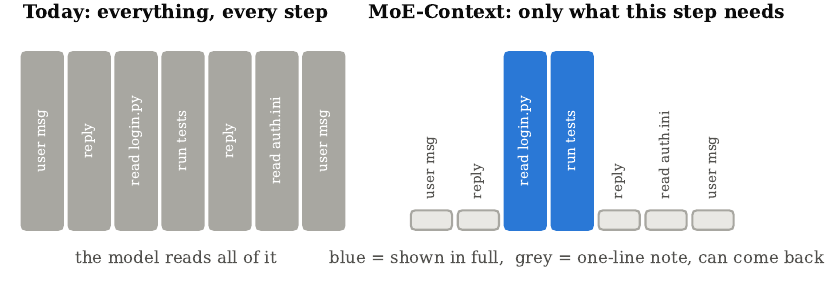}
\caption{Left: today the model reads everything at every step. Right: MoE-Context shows only the blocks this step needs; the rest stay as one-line notes.}
\label{fig:idea}
\end{figure}

MoE-Context treats the context like memory (Figure~\ref{fig:idea}). The context is a set of \textbf{blocks}: a message, a reply, a tool call with its result, a file read. Before each step a fast \textbf{decide pass} asks of every block: \emph{is this needed for the next step?} Needed blocks are shown in full. The rest are shown as a \textbf{one-line note}, such as ``read login.py, 300 lines''. If the model misses something it \textbf{recalls} the block; nothing is deleted. Some blocks, like instructions, are \textbf{pinned}.

\begin{figure}[htbp]
\centering
\includegraphics[width=0.92\linewidth]{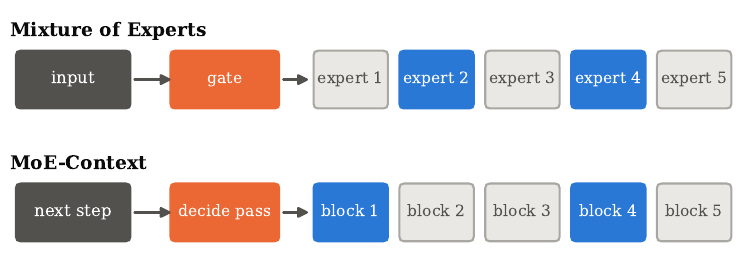}
\caption{Mixture of Experts picks experts for each input. MoE-Context picks context blocks for each step.}
\label{fig:moe}
\end{figure}

The name comes from Mixture of Experts~\citep{jacobs1991,shazeer2017}, where a gate picks a few experts per input; here the decide pass picks a few blocks per step (Figure~\ref{fig:moe}). The decide pass plays the fast ``System 1'' role and the model that reasons and writes plays the slow ``System 2'' role~\citep{kahneman2011}.

\paragraph{Contributions.}
\begin{enumerate}[leftmargin=*]
\item A design for per-step, reversible block switching with notes, pinning and recall (Section~\ref{sec:method}).
\item A single-model version that decides and answers over one key-value cache with attention masking, so blocks are read once and never re-read, with a laptop implementation (Sections~\ref{sec:method} and~\ref{sec:exp}).
\item Evidence that less in view helps a small model, and that masking hides content as well as deletion (Section~\ref{sec:exp}).
\item An honest account of where the decide pass falls short: a keyword baseline beats it on our task, and a supervised linear probe that sharpens it does not transfer (Section~\ref{sec:exp}).
\item Two usage patterns that block switching makes cheap: undoing a failed attempt and subagents without a new context (Section~\ref{sec:patterns}).
\end{enumerate}

\section{Related work}\label{sec:related}

Many systems shorten an agent's context. Most cut or summarize it, and the original is gone~\citep{xiao2025agentdiet,kang2025acon,wu2025resum,yi2026adacom,kontonis2026memento}. Some keep the original and let the model fetch it back, but the model has to stop and manage its context itself~\citep{packer2023memgpt,li2026sculptor,li2026acm,busch2026ctxagent,su2026refold}, or fixed rules decide what to show~\citep{mason2026pichay,ruan2026folio}. Closest to our decide pass, winnow scores chunks of each tool result once with a decision model~\citep{winnow2026}, SWE-Pruner Pro reads keep-or-prune decisions per line from a coding agent's own hidden states through a small trained head and deletes pruned lines~\citep{wang2026sweprunerpro}, and Neural Paging proposes a learned controller without testing it on a real model~\citep{chen2026neuralpaging}. Closest to our single-cache version, Cache Engineering lets the agent or a fixed rule drop key-value entries from future attention; its engine can rebuild evicted cache by recomputing it, but it gives no way to return a dropped entry to attention~\citep{fu2026cache}. Jev-Mem pairs a fast decision model with a slower answering model for long-term memory~\citep{jiang2026jevmem}. Table~\ref{tab:compare} summarizes.

\begin{table}[htbp]
\centering
\small
\caption{What each approach does (\yes{} yes, \some{} partly).}
\label{tab:compare}
\begin{tabular}{lcccc}
\toprule
 & Keeps the & Can bring & Checks every & One model \\
 & original & it back & block, every step & picks and answers \\
\midrule
Cut or summarize~\citep{xiao2025agentdiet,kang2025acon,kontonis2026memento} & & & & \\
Model manages itself~\citep{packer2023memgpt,li2026sculptor,su2026refold} & \yes & \yes & & \\
Fixed rules~\citep{mason2026pichay,ruan2026folio} & \yes & \yes & \some & \\
Learned controller, untested~\citep{chen2026neuralpaging} & \yes & \yes & \yes & \\
Decision model, once~\citep{winnow2026} & \yes & \yes & & \\
Own hidden states, once~\citep{wang2026sweprunerpro} & & & & \some \\
Cache drop~\citep{fu2026cache} & & & & \some \\
\textbf{MoE-Context} & \yes & \yes & \yes & \yes \\
\bottomrule
\end{tabular}
\end{table}

\section{Method}\label{sec:method}

\begin{figure}[htbp]
\centering
\includegraphics[width=0.95\linewidth]{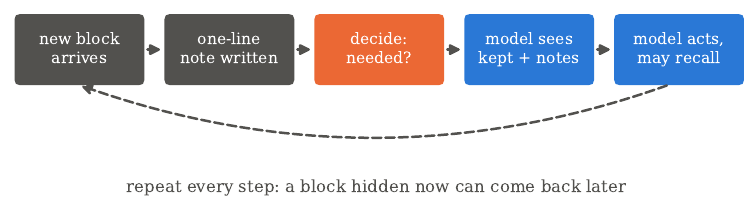}
\caption{The loop. The decide pass writes no text; it scores each block.}
\label{fig:loop}
\end{figure}

\paragraph{The loop.} On every step (Figure~\ref{fig:loop}): a new block arrives and the model writes its one-line note; the decide pass scores every block; the context is built from pinned blocks, kept blocks and the notes of the rest; the model acts and may recall a block. A block hidden now can return later.

\paragraph{Why the decide pass reads full blocks.} A note cannot hold every detail. If a bug sits on line 140 of a 300-line file, the note ``read login.py'' will not show it. The decide pass therefore reads full blocks; notes exist so the answering model knows what is available.

\begin{figure}[htbp]
\centering
\includegraphics[width=0.95\linewidth]{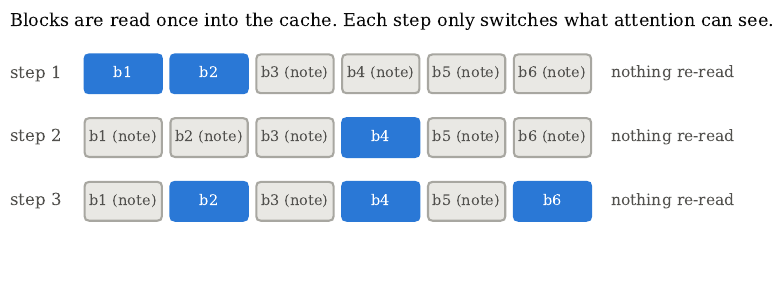}
\caption{One model, one cache. Each step only switches which blocks attention can see.}
\label{fig:cache}
\end{figure}

\paragraph{Two ways to build it.} With \emph{two models}, a small decision model such as Jev~\citep{typesafe2026} picks blocks for a large model; we did not test this version. With \emph{one model}, the same model first scores each block, then answers. Each block is read once into the key-value cache (the stored result of reading each token). Hiding a block sets its entries to zero in the attention mask; positions are kept, so cached entries stay valid and nothing is re-read (Figure~\ref{fig:cache}). Each block has a short header with its note, and a tag shown only while it is hidden.

\paragraph{Cache order.} Blocks are encoded once, in order, over the full sequence. A later block has therefore already attended to earlier blocks, including ones that are later hidden, so traces of a hidden block can remain inside later blocks. We test this in Section~\ref{sec:exp}.

\paragraph{The decide pass.} For block $i$ and question $q$ the model sees only block $i$ and the prompt ``Is the content of block $i$ needed to answer this question? Answer yes or no.'' We read the last-position output scores instead of generating text.

\paragraph{A supervised linear probe written into the weights.} The raw yes/no scores of a small model are poorly separated. We fit a supervised linear probe on labelled practice questions, which is Fisher's linear discriminant on the final hidden state $h$:
\begin{equation}
w = (\Sigma + \lambda I)^{-1}(\mu_{\text{needed}} - \mu_{\text{not needed}}),
\end{equation}
where $\mu$ are class means, $\Sigma$ the shared covariance and $\lambda$ a ridge term. We remove $w$'s component along the mean hidden state and fit a scale and offset on held-out data. This is training, but with a closed-form fit and no gradient updates to the model. We then write the probe into two spare output rows that the tokenizer never uses; the decide pass reads them and the answer step forbids them, so normal answers cannot change. Prior work shows behaviours can be read as directions in hidden states and added to activations or written into weights~\citep{arditi2024refusal,panickssery2023caa}. Writing into the real ``yes'' and ``no'' rows instead broke ordinary answers in our first attempts.

\section{Experiments}\label{sec:exp}

\paragraph{Setup.} Qwen3-1.7B (bf16, thinking mode off, greedy decoding) on an 8\,GB laptop GPU. Each layout has 16 blocks: 8 settings files of 30 key and value lines and 8 test logs, about 3,450 tokens. Three layouts with 40 questions each give 120 questions; each asks for one value in one file. The probe's practice data uses file names and keys that never appear in the test. All settings are listed in Appendix~\ref{app:repro}.

\begin{figure}[htbp]
\centering
\includegraphics[width=0.85\linewidth]{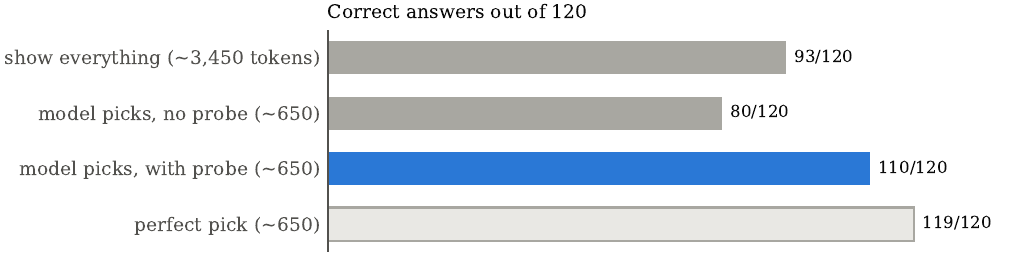}
\caption{Correct answers out of 120. Token counts are what the answer step sees.}
\label{fig:main}
\end{figure}

\paragraph{Showing less helps.} With a plain prompt the model answered 93 of 120 correctly when shown everything and 119 of 120 when shown only the needed block, about five times fewer tokens (Figure~\ref{fig:main}).

\begin{figure}[htbp]
\centering
\includegraphics[width=0.9\linewidth]{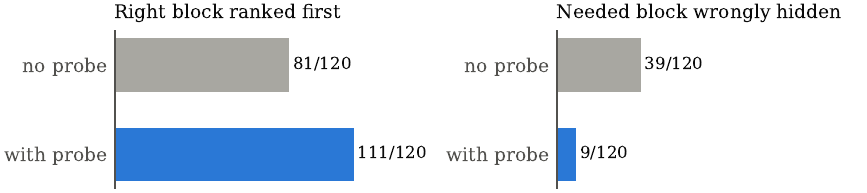}
\caption{The probe, out of 120 questions.}
\label{fig:edit}
\end{figure}

\paragraph{The model's own picking, with and without the probe.} The decide pass ranked the needed block first 81 of 120 times without the probe and 111 with it (Figure~\ref{fig:edit}). Picking the top block and answering scored 80 of 120 without the probe, worse than showing everything, and 110 of 120 with it.

\begin{figure}[htbp]
\centering
\includegraphics[width=0.95\linewidth]{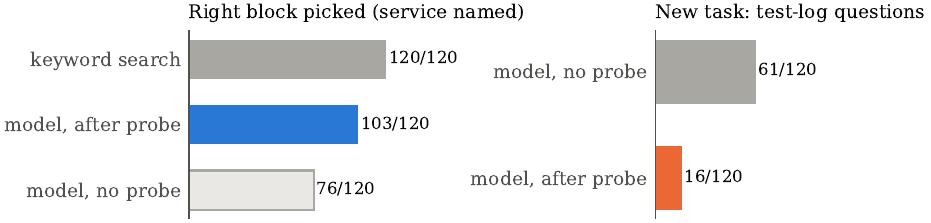}
\caption{Reviewer checks. Left: block picked correctly when questions name the service but not the file. Right: the probe on a task it was not fitted on.}
\label{fig:checks}
\end{figure}

\paragraph{A keyword baseline beats the decide pass on this task.} Each question names its file, so we added the baseline a reviewer suggested: pick the block with the most word overlap with the question. It picked the right block 120 of 120 times and answered 119 of 120. With questions that name the service in words but not the file name, keyword search still picked 120 of 120, while the decide pass picked 76 of 120 without the probe and 103 with it (Figure~\ref{fig:checks}, left). On this task, simple retrieval is enough; the decide pass is not yet justified by accuracy alone.

\paragraph{The probe does not transfer.} We asked a second kind of question, about one line in a test log. Without the probe the decide pass ranked the needed block first 61 of 120 times; with the probe, fitted only on settings-file questions, it fell to 16 of 120, usually picking a settings file instead (Figure~\ref{fig:checks}, right). The probe learned the training task, not relevance in general.

\begin{figure}[htbp]
\centering
\includegraphics[width=0.85\linewidth]{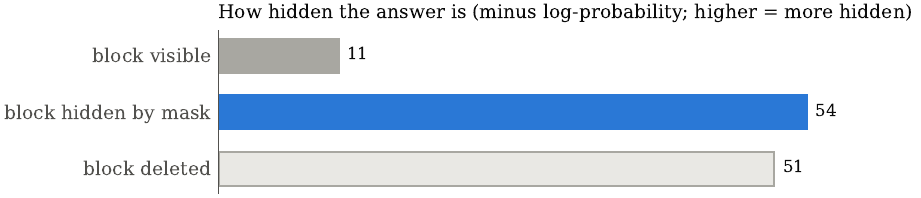}
\caption{Averaged over 120 questions. A masked block is as hidden as a deleted one.}
\label{fig:leak}
\end{figure}

\paragraph{Hidden means hidden.} We compared how likely the model is to produce the hidden answer when the block is visible (log-probability about $-11$), hidden by the mask (about $-54$), and removed from the input with a fresh encode (about $-51$). Masking hid content as well as deletion here (Figure~\ref{fig:leak}). In this layout the needed block comes before most other blocks, so this test does not cover traces carried by later blocks; see the undo test below.

\paragraph{Recall prompting fails for a small model.} When the prompt explained recall and tagged hidden blocks, the model asked for blocks far too often, even with the answer in view, and accuracy fell to 44 of 120 with everything visible. When only the needed block was hidden and it asked, it named that block 72 of 72 times. Asking at the right time needs a stronger model or a learned skill.

\section{Two usage patterns}\label{sec:patterns}

\begin{figure}[htbp]
\centering
\includegraphics[width=0.9\linewidth]{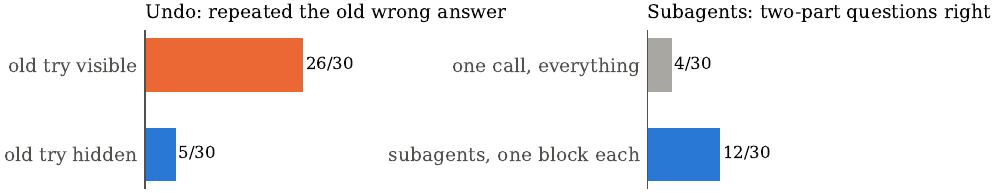}
\caption{Left: hiding a failed attempt mostly stops the model from repeating it. Right: subagents that each see one block beat one call that sees everything. 30 questions each.}
\label{fig:patterns}
\end{figure}

Hiding and showing a block is only a mask change, so two patterns become cheap (Figure~\ref{fig:patterns}).

\paragraph{Undo a failed attempt.} Switch off the blocks of a failed attempt and retry with a clean view; the attempt stays recallable. With an earlier wrong attempt visible, the model copied its wrong value 26 of 30 times and answered correctly 3 times. With it hidden, it copied the wrong value 5 times and answered correctly 16 times. We traced those 5 copies. In 13 of the 30 cases the model-written note for the failed attempt contained the old wrong value, and all 5 copies came from those cases (5 of 13; 0 of 17 otherwise). Hiding the note as well as the body removed every copy (0 of 30) while keeping 15 of 30 correct. The leak is the note, not the masked cache, so notes for failed attempts should not repeat their wrong outputs.

\paragraph{Subagents without a new context.} For a side task, switch on only the blocks it needs, keep its short result, and switch the rest back. On two-part questions, one call that saw everything answered 4 of 30; lookups that each saw one file answered 12 of 30, limited by single-lookup accuracy (38 of 60). Sculptor lists suppressing erroneous steps as future work~\citep{li2026sculptor}, and Pichay reports shared subagent state as a problem~\citep{mason2026pichay}.

\section{Discussion}\label{sec:discussion}

\paragraph{What holds.} Less in view helps this model; masking a cached block hides it as well as deleting it; and block switching makes undo and subagents cheap without re-reading.

\paragraph{What does not hold yet.} The decide pass is not better than keyword search on a task where the question names its target, and a probe fitted on one task hurts on another. A fair test needs tasks where relevance is not visible in the words, such as finding the file that caused a failing test, and a decide pass trained on many task types.

\paragraph{Cost.} Changing which blocks are shown can break the discount providers give for repeated prompts. With $N$ tokens in all blocks, $A$ shown in full, $S$ in notes, and prices $p_{\text{in}}$ (input), $p_{\text{cache}}$ (cached input) and $p_c$ (decide pass), MoE-Context is cheaper when
\begin{equation}
\frac{A+S}{N} < \frac{p_{\text{cache}} - p_c}{p_{\text{in}}}.
\end{equation}
If cached input costs a tenth of normal input, the shown part must stay under about a tenth of everything, so the main argument is quality, not price. The single-cache version avoids this problem.

\paragraph{Limits.} One small model and one synthetic task family; 120 questions without error bars; the decide pass runs one forward pass per block (about one second per question here), so we make no speed claim; blocks are judged one at a time; a poor note can make a block hard to find; serving stacks and public APIs do not yet expose attention masking.

\section{Conclusion and future work}

MoE-Context shows a model only the blocks a step needs, keeps everything else recallable, and in its single-model form does so over one cache without re-reading. Showing less helped, masking hid content as well as deletion, and block switching made undo and subagents cheap. The model's own picking is not yet better than keyword search, and our probe did not generalize.

\begin{figure}[htbp]
\centering
\includegraphics[width=\linewidth]{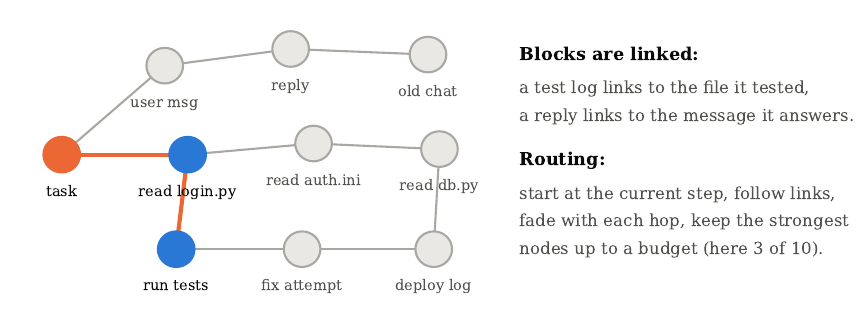}
\caption{Future work: blocks as a graph. Routing starts at the current step, spreads along links, and stops at a budget.}
\label{fig:graph}
\end{figure}

\paragraph{Blocks as a graph.} Blocks are connected: a test log belongs to the file it tested, a reply to the message it answers. Storing these links turns picking into routing (Figure~\ref{fig:graph}): start at the current step, follow links outward with each hop counting less, and keep the strongest blocks up to a budget, for example 30\% of all blocks. This could avoid keeping duplicate blocks and pull in blocks that look irrelevant alone.

\paragraph{Next steps.} Harder tasks where keyword search fails; a decide pass trained on many task types; comparison with embedding retrievers; training the picker and the model as a pair; larger models; and serving support for switching blocks on and off.

\paragraph{Acknowledgements.} We thank Matthias Busch for endorsing this work and for detailed suggestions that led to the keyword baseline, the transfer test and the cache-order analysis in this version.

\appendix
\section{Reproducibility details}\label{app:repro}

\begin{itemize}[leftmargin=*]
\item \textbf{Model and decoding.} Qwen3-1.7B in bf16; chat template with thinking disabled everywhere; greedy decoding, at most 24 new tokens.
\item \textbf{Block layout in the cache.} Each block is ``\texttt{[block $i$ | note]}'', then a tag ``\texttt{(hidden)}'' visible only while the body is hidden, then the body and ``\texttt{[/block $i$]}''. Hidden bodies are masked out of attention with explicit position ids.
\item \textbf{Prompts.} Note: ``Describe this context block in under 15 words, one line. Name the tool and any file name it mentions.'' Decide question: ``Question: \{q\}. Is the content of block $i$ needed to answer this question? Answer yes or no.'', with only block $i$'s body visible. Plain answering: system prompt ``Answer the question using the context.'' with only the chosen block texts. Recall runs add: ``If you need the content of a (hidden) block, reply exactly: RECALL $<$block number$>$.''
\item \textbf{Questions.} ``What is the value of \{key\} in config/\{name\}.ini? Answer with the value only.'' Paraphrased: ``What is the \{key\} setting for the \{name\} service?'' Test-log task: whether a named test passed or was skipped.
\item \textbf{Decide score.} Sigmoid of the difference between the log-sum-exp of ``yes'' and ``no'' token scores at the last position.
\item \textbf{Probe.} Final hidden state (input to the output layer) at the last token. $\lambda = 0.1 \times$ the mean of the diagonal of $\Sigma$; component along the mean hidden state removed; scaled to unit spread. Scale $a = 4.691$ and offset $c = -9.030$ fitted with L-BFGS on binary cross-entropy. Written as $\pm\tfrac{a}{2}w$ into output rows 151869 and 151870 (tokenizer size plus 200 and 201), which the answer step forbids. The model's input and output embeddings are tied in its configuration; the edit is applied only after checking they are separate tensors in memory.
\item \textbf{Data sizes and seeds.} Probe fit: seeds 101 to 110, 2,400 samples (150 positive). Calibration: seeds 201 to 204, 960 samples (60 positive). Fit keys and file names never appear in the test. Test layouts: seeds 0, 1, 2 with 40 questions each. Undo test: seed 0; subagent test: seed 1000.
\item \textbf{Code.} The scripts and raw outputs are available from the author on request.
\end{itemize}

\end{document}